\pdfoutput=1

\documentclass[11pt]{article}

\usepackage{ACL2023}

\usepackage[utf8]{inputenc}   
\usepackage[T1]{fontenc}
\DeclareUnicodeCharacter{202F}{\nobreakspace}

\usepackage{times}
\usepackage{latexsym}
\usepackage{amssymb}
\usepackage{graphicx}

\usepackage{booktabs}        
\usepackage[table]{xcolor}   
\usepackage{array}           
\usepackage{multirow}        

\usepackage{amsmath}

\usepackage[T1]{fontenc}

\usepackage[utf8]{inputenc}

\usepackage{microtype}

\usepackage{inconsolata}

\usepackage{algorithm}
\usepackage{algorithmicx}
\usepackage{algpseudocode}

\title{MMoA: An AI-Agent framework with recurrence for Memoried Mixure-of-Agent}

\author{Rui Chu}

\begin{document}
\maketitle
\begin{abstract}
The Mixture-of-Agents (MoA) framework has shown promise in enhancing large language model (LLM) performance by aggregating outputs from multiple agents. However, existing MoA systems rely on static routers that neglect temporal and contextual dependencies across aggregation layers. To address this limitation, we propose \textbf{MMoA}, a novel recurrent MoA architecture that integrates LSTM-based gating into the agent selection process. Our recurrence router adaptively modulates agent contributions based on both current inputs and historical routing decisions, enabling more informed and context-aware aggregation. We evaluate MMoA across standard instruction-following benchmarks including AlpacaEval 2.0, MT-Bench, and Arena-Hard. Results show that MMoA achieves comparable accuracy to traditional MoA while significantly reducing computational overhead by dynamically activating fewer agents. For instance, on AlpacaEval 2.0, MMoA achieves a win rate of 58.0\% with only a minor drop compared to MoA's 59.8\%, while offering up to 4.6\% improvement in runtime efficiency. Our approach offers a scalable and efficient solution for adaptive multi-agent LLM systems.

\end{abstract}

\section{Introduction}

Recent advancements in large language models (LLMs) have demonstrated that collaborative architectures can significantly enhance performance on complex language understanding and generation tasks. In particular, the Mixture-of-Agents (MoA) framework leverages multiple LLM agents to iteratively refine responses through an aggregation process. This design harnesses the complementary strengths of diverse models, leading to improved accuracy and robustness over individual agents.

Despite its success, the standard MoA approach typically employs a static routing mechanism that aggregates agent outputs without explicitly considering the temporal or contextual dependencies between successive aggregation layers. In contrast, recent developments in the Mixture-of-Experts (MoE) domain have explored adaptive gating mechanisms \cite{li2023adaptive} and layerwise recurrent routers \cite{qiu2024layerwise} that incorporate dynamic selection strategies to adjust expert contributions based on input complexity and historical routing decisions. Such mechanisms, often implemented via LSTM or RNN architectures, have proven effective in balancing computational efficiency with model performance by modulating the flow of information across layers.

Motivated by these insights, we propose a novel integration of a recurrent gating mechanism into the MoA framework. Our approach embeds an LSTM/RNN-based gating module within the routing process, enabling the model to dynamically re-weight and fuse the outputs of multiple agents. By leveraging both current input features and historical routing information, the recurrent gating module adaptively adjusts the contribution of each agent, thereby enhancing the overall aggregation quality while preserving the beneficial diversity inherent in a multi-agent system.

The key contributions of this work are as follows:
\begin{itemize}
    \item \textbf{Router Design for MoA Agent Selection:} We propose a router adapted from MoE’s LSTM/RNN-based mechanisms to perform dynamic, context-aware agent selection within the MoA framework, efficiently leveraging past routing decisions to modulate agent contributions.
    \item \textbf{Recurrent Gating Integration:} We introduce a recurrent gating mechanism that combines the strengths of LSTM/RNN architectures with the MoA framework, allowing for context-aware and temporally informed routing decisions.
\end{itemize}
To the best of our knowledge, this is the first work to introduce a router for agent selection in the MoA framework, thereby significantly enhancing aggregation efficiency and overall model performance. With our method, we can first-time Tune the Mixture of Agent system in a Language Model way.

Through this work, we aim to bridge the gap between static aggregation strategies and dynamic, recurrent routing mechanisms, thereby paving the way for more intelligent and adaptive multi-agent LLM systems in real-world applications.

\begin{figure}
    \centering
    \includegraphics[width=1\linewidth]{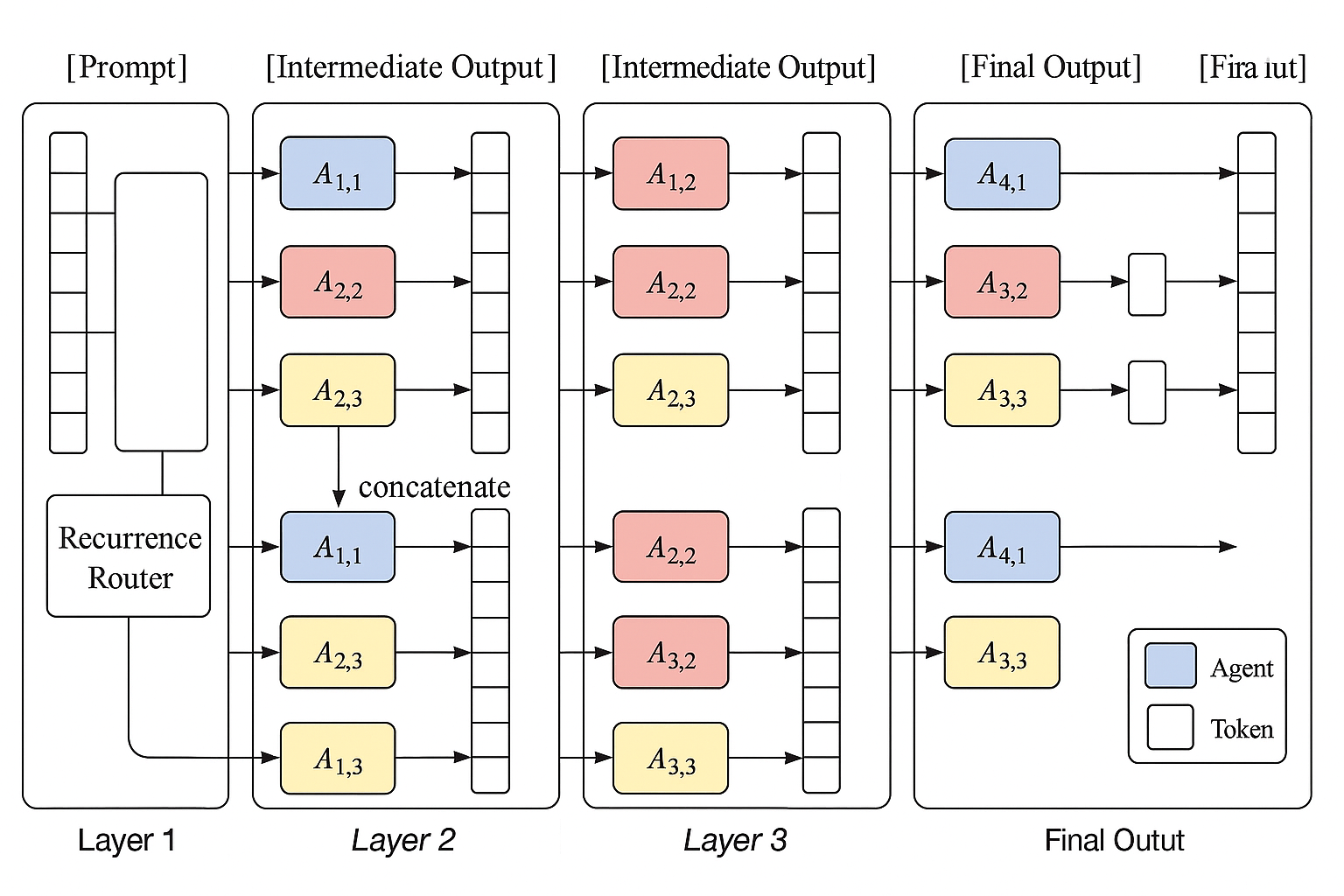}
    \caption{Effectiveness of Recurrence Router in MoA system: The router compared to the upper section which is the original MoA, the recurrence router (a Recurence architecture of neural network) in MMoA can find out the most valuable agents in the following layers}
    \label{fig:mainFig}
\end{figure}

\section{Related Works}

\textbf{Mixture of Agents.} The Mixture-of-Agents (MoA) framework has emerged as a promising approach to enhance large language model (LLM) capabilities by leveraging the collaborative strengths of multiple agents. In this paradigm, several independent agents generate candidate responses which are then aggregated by a dedicated router to produce a final output. This modular design not only enables the synthesis of diverse perspectives and complementary expertise but also mitigates individual model limitations by effectively combining outputs. The MoA framework has demonstrated significant improvements in performance across various benchmarks, thereby establishing itself as a viable architecture for scalable and robust language understanding.
\textbf{Router Design in Mixture of Experts.} The Mixture-of-Experts (MoE) framework relies critically on effective router designs to assign input tokens to specialized expert modules. Pioneering work by Shazeer et al. (2017) introduced the Sparsely-Gated MoE, employing a linear gating layer with Softmax and Top-$k$ sparse activation to reduce computational costs, alongside load-balancing regularizers to prevent expert overuse \cite{shazeer2017outrageously}. Subsequent advancements, such as the Switch Transformer, simplified routing to Top-1 expert selection with noisy gating to encourage exploration. Other systems optimized large-scale deployments with lightweight Top-$k$ routing and sharding to reduce cross-device communication. More recent work has explored expert-choice style routing and richer token--expert interaction mechanisms. These evolving router designs balance computational efficiency, scalability, and expert utilization, significantly boosting MoE model performance.
\textbf{Recurrent Mechanisms in Mixture of Experts.} Complementing the MoA framework, recent advances in the Mixture-of-Experts (MoE) domain have focused on integrating recurrent mechanisms to enhance the router’s decision-making process. Adaptive gating approaches have incorporated LSTM-based modules to dynamically adjust the number of experts processing each token, thus tailoring computational efforts according to the input's complexity \cite{li2023adaptive}. In parallel, layerwise recurrent routers utilize RNN architectures to propagate historical routing information across layers, enabling the router to make more informed and consistent expert selections \cite{qiu2024layerwise}. These recurrent strategies enrich the routing process with memory and gating information, resulting in a flexible and context-aware mechanism that not only improves computational efficiency but also enhances overall model performance.

\section{Method}

In the standard MoA framework, multiple independent agents generate candidate responses for a given input, and a static router aggregates these responses without considering temporal dependencies. In contrast, our method introduces a recurrent gating module—implemented via an LSTM (or RNN)—that processes both the outputs of the agents and the hidden state from previous aggregation layers. This recurrent structure enables the model to adaptively fuse information across layers.

\subsection{Recurrent Gating Module}
Let $\{A_1(x), A_2(x), \ldots, A_n(x)\}$ denote the outputs of $n$ agents for an input $x$. Instead of aggregating these outputs statically, we first fuse them into a single representation:
\begin{equation}
    z_t = f\big(A_1(x), A_2(x), \ldots, A_n(x)\big),
\end{equation}
where $f(\cdot)$ is a fusion function, such as concatenation followed by a linear projection.

The fused representation $z_t$ is then processed by an LSTM to capture historical context:
\begin{equation}
    h_t = \mathrm{LSTM}(z_t, h_{t-1}),
\end{equation}
where $h_{t-1}$ is the hidden state from the previous aggregation layer and $h_t$ is the updated hidden state.

Next, we compute a gating vector $g \in \mathbb{R}^n$ using a linear transformation followed by a softmax activation:
\begin{equation}
    g = \mathrm{softmax}(W_g h_t + b_g),
\end{equation}
with $W_g$ and $b_g$ as learnable parameters. The gating vector $g$ is then used to dynamically weight the outputs of the agents:
\begin{equation}
    y = \sum_{i=1}^{n} g_i \cdot A_i(x).
\end{equation}

\subsection{Integration with the MoA Framework}
The recurrent gating module is seamlessly integrated into the multi-layer MoA architecture. At each layer $l$, the module receives the candidate responses $\{A_i^{(l)}(x)\}_{i=1}^{n}$ and the aggregated output from the previous layer, $y^{(l-1)}$. The aggregation process at layer $l$ is formulated as:
\begin{equation}
    y^{(l)} = G\Big(y^{(l-1)}, \{A_i^{(l)}(x)\}_{i=1}^{n}\Big),
\end{equation}
where $G(\cdot)$ denotes the recurrent gating module. This design allows the model to iteratively refine its aggregated output, leveraging both current input features and historical routing information.

The detailed steps can be refered to Algorithm \ref{alg:rgmoa}.


\begin{algorithm}[htbp]
\caption{Recurrent Gating Mixture-of-Agents }
\label{alg:rgmoa}
\begin{algorithmic}[1]
\State \textbf{Input:} Input prompt $x$, number of layers $L$, number of agents $n$, agent functions $\{A_1, A_2, \dots, A_n\}$, fusion function $f(\cdot)$, LSTM parameters, gating parameters $W_g$, $b_g$, initial hidden state $h_0$
\State \textbf{Output:} Final aggregated output $y^{(L)}$
\State $x^{(0)} \gets x$
\For{$l = 1$ to $L$}
    \For{$i = 1$ to $n$} 
        \State $a_i^{(l)} \gets A_i^{(l)}(x^{(l-1)})$
    \EndFor
    \State $z^{(l)} \gets f\big(a_1^{(l)}, a_2^{(l)}, \dots, a_n^{(l)}\big)$ 
    \State $h^{(l)} \gets \mathrm{LSTM}\big(z^{(l)}, h^{(l-1)}\big)$
    \State $g^{(l)} \gets \mathrm{softmax}\big(W_g\, h^{(l)} + b_g\big)$ 
    \State $y^{(l)} \gets \sum_{i=1}^{n} g^{(l)}_i \cdot a_i^{(l)}$ 
    \State $x^{(l)} \gets y^{(l)}$
\EndFor
\State \Return $y^{(L)}$
\end{algorithmic}
\end{algorithm}

\subsection{Router Loss Function for Agent Selection}

To train a Router for selecting optimal Agents in a Mixture of Agents (MoA) framework, we define a loss function that encourages the Router to assign tasks to Agents that minimize task-specific losses while maintaining exploration through regularization. Let \( x \) denote the input task, \( A_i \) the \( i \)-th Agent, and \( y_i \) the output of \( A_i \) for \( x \). The target output is \( y^* \), and the task-specific loss for \( A_i \) is \( L_i = L(y_i, y^*) \).

The Router outputs a probability distribution over \( N \) Agents:
\[
p = [p_1, p_2, \dots, p_N], \quad \sum_{i=1}^N p_i = 1,
\]
where \( p_i \) is the probability of selecting Agent \( A_i \).

The loss function for the Router is defined as:
\[
L_{\text{total}} = \sum_{i=1}^N p_i L_i + \lambda H(p),
\]
where:
\begin{itemize}
    \item \( \sum_{i=1}^N p_i L_i \): The expected task loss, weighting each Agent's loss by its selection probability.
    \item \( H(p) = -\sum_{i=1}^N p_i \log p_i \): The entropy of the Router's probability distribution, encouraging exploration.
    \item \( \lambda \): A hyperparameter controlling the strength of entropy regularization. 
\end{itemize}

Optionally, to promote load balancing across Agents, a load balancing term can be added:
\[
L_{\text{load}} = \sum_{i=1}^N \left( p_i - \frac{1}{N} \right)^2.
\]
The extended loss becomes:
\[
L_{\text{total}} = \sum_{i=1}^N p_i L_i + \lambda H(p) + \gamma L_{\text{load}},
\]
where \( \gamma \) is a hyperparameter for load balancing.

This loss function ensures the Router learns to select Agents that minimize task losses while maintaining a balanced and exploratory selection strategy.

\section{Experiment}
\subsection{Settings and Data}

In our experiments, we follow a standard instruction-following evaluation setup to ensure a fair and comprehensive comparison of our proposed method. We evaluate on the following benchmark datasets:

\begin{itemize}
    \item \textbf{AlpacaEval 2.0:} This benchmark consists of 805 instructions representative of real-world applications. Each model's response is evaluated using a length-controlled (LC) win rate metric to eliminate biases due to output length.
    \item \textbf{Arena-Hard:} A challenging dataset comprising 500 diverse queries that span a range of topics, including coding, mathematics, and logic puzzles. This benchmark tests the robustness and versatility of the models.
    \item \textbf{MT-Bench:} In this multi-turn evaluation framework, the responses are scored (typically on a scale from 0 to 10) using a GPT-4-based evaluator, offering a fine-grained assessment of model performance.
    \item \textbf{FLASK:} A benchmark that provides detailed evaluations across 12 skill-specific dimensions, further highlighting the strengths and weaknesses of the models.
\end{itemize}

While the primary focus of our work is on enhancing the inference-time aggregation mechanism via a recurrent gating module, our overall training and evaluation pipeline follows common practices in instruction-tuned LLM development, ensuring that observed gains are attributable to the proposed recurrent routing mechanism.



For comparison, we consider the following baselines:
\begin{itemize}
    \item \textbf{Baseline MoA:} The standard MoA framework with a static aggregation mechanism.
    \item \textbf{Ablated Model:} A variant where the recurrent gating module is replaced by a simple linear layer, thereby removing the temporal dependency component.
\end{itemize}

\subsubsection{Performances on Accuracy}

\begin{figure}
    \centering
    \includegraphics[width=1\linewidth]{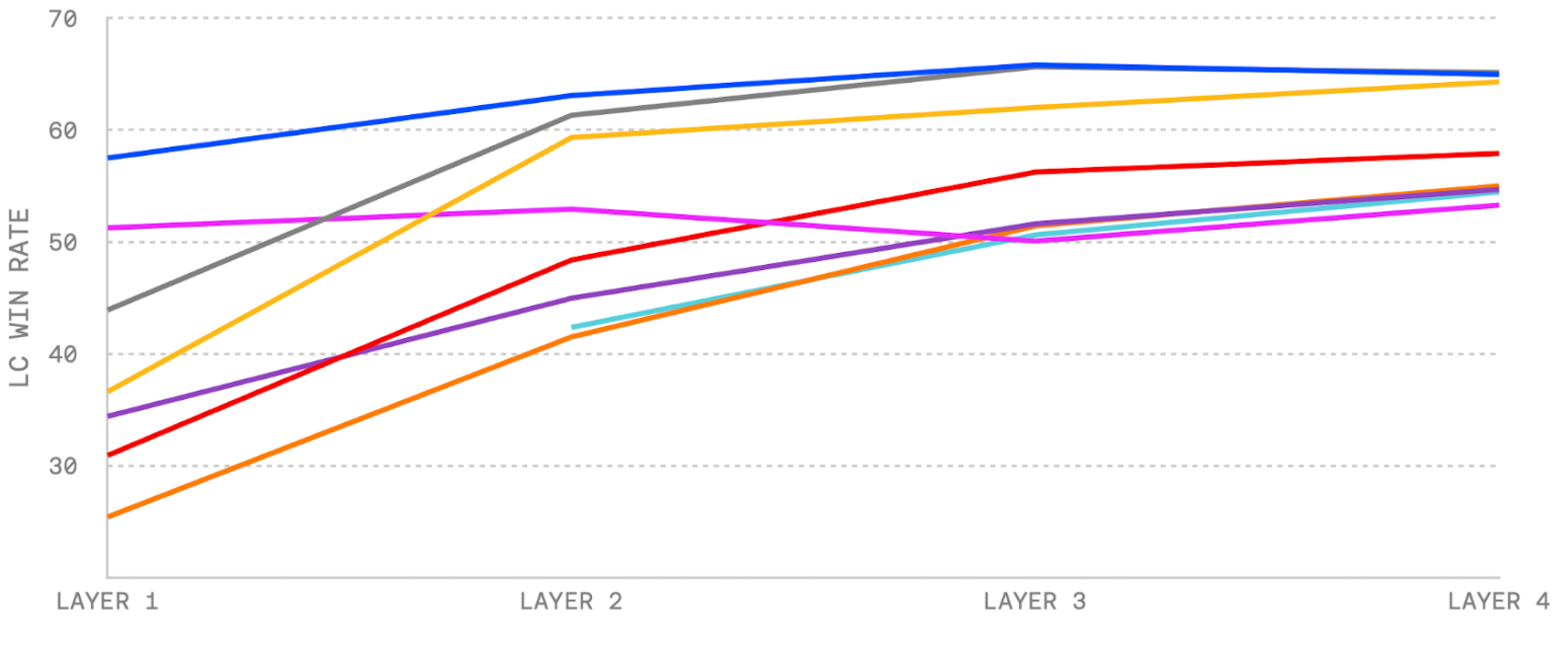}
    \caption{Blue Line shows the accuracy of MMoA}
    \label{fig:mmoa_performance}
\end{figure}
As shown in Figure~\ref{fig:mmoa_performance}, beginning at \textbf{Layer 2}, the sky-blue MMoA curve jumps to about 42\% LC win rate—significantly outperforming the original MoA at that depth. Although its win rate at \textbf{Layers 3} and \textbf{4} (approximately 50\% and 55\%, respectively) falls slightly behind the top performers, this modest late-layer drop is offset by the fact that MMoA activates far fewer agents overall. In other words, MMoA achieves very high accuracy early on while striking an excellent balance between agent count (and thus computational cost) and final win rate as network depth increases.

\begin{table}[htbp]
  \centering
  \caption{(a) AlpacaEval 2.0}
  \label{tab:alpacaeval-narrow}
  \small
  \resizebox{\columnwidth}{!}{%
    \begin{tabular}{@{}lcc@{}}
      \toprule
      Model               & LC win.           & win.             \\
      \midrule
      \rowcolor{gray!20} MoA w/ GPT-4o    & $65.7\pm0.7\%$    & $78.7\pm0.2\%$   \\
      \rowcolor{gray!20} MoA                & $65.1\pm0.6\%$    & $59.8\pm0.3\%$   \\
      \rowcolor{gray!50} \textbf{MMoA}      & $\mathbf{61.5\pm0.4\%}$ & $\mathbf{58.0\pm0.5\%}$ \\
      \rowcolor{gray!20} MoA-Lite           & $59.3\pm0.2\%$    & $57.0\pm0.7\%$   \\
      GPT-4 Omni (05/13)   & 57.5\%            & 51.3\%           \\
      GPT-4 Turbo (04/09)  & 55.0\%            & 46.1\%           \\
      WizardLM 8×22B\textsuperscript{\dag} & 51.3\%            & 62.3\%           \\
      GPT-4 Preview (11/06)& 50.0\%            & 50.0\%           \\
      Qwen1.5 110B Chat    & 43.9\%            & 33.8\%           \\
      Qwen1.5 72B Chat     & 36.6\%            & 26.5\%           \\
      GPT-4 (03/14)        & 35.3\%            & 22.1\%           \\
      Llama 3 70B Instruct & 34.4\%            & 33.2\%           \\
      Mixtral 8×22B v0.1   & 30.9\%            & 22.2\%           \\
      \bottomrule
    \end{tabular}%
  }
\end{table}

\begin{table}[htbp]
  \centering
  \caption{(b) MT-Bench}
  \label{tab:mtbench-narrow}
  \small
  \resizebox{\columnwidth}{!}{%
    \begin{tabular}{@{}lccc@{}}
      \toprule
      Model                      & Avg.              & 1st turn         & 2nd turn         \\
      \midrule
      \rowcolor{gray!20} MoA w/ GPT-4o    & $9.40\pm0.06$     & 9.49             & 9.31             \\
      GPT-4 Turbo (04/09)        & 9.31              & 9.35             & 9.28             \\
      \rowcolor{gray!20} MoA                  & $9.25\pm0.10$     & 9.44             & 9.07             \\
      \rowcolor{gray!50} \textbf{MMoA}        & $\mathbf{9.20\pm0.08}$  & $\mathbf{9.42}$ & $\mathbf{9.05}$  \\
      GPT-4 Preview (11/06)      & 9.20              & 9.38             & 9.03             \\
      GPT-4 Omni (05/13)         & 9.19              & 9.31             & 9.07             \\
      \rowcolor{gray!20} MoA-Lite             & $9.18\pm0.09$     & 9.38             & 8.99             \\
      Qwen1.5 110B Chat          & 8.96              & 9.23             & 8.63             \\
      Llama 3 70B Instruct       & 8.94              & 9.20             & 8.68             \\
      Mixtral 8×22B v0.1         & 8.78              & 9.11             & 8.44             \\
      WizardLM 8×22B             & 8.78              & 8.96             & 8.61             \\
      Qwen1.5 72B Chat           & 8.44              & 8.55             & 8.34             \\
      GPT-4 (06/13)              & 8.84              & 9.08             & 8.61             \\
      \bottomrule
    \end{tabular}%
  }
\end{table}

\paragraph{Analysis.}
In both AlpacaEval 2.0 (Table~\ref{tab:alpacaeval-narrow}) and MT-Bench (Table~\ref{tab:mtbench-narrow}), MMoA (dark‐shaded row) shows only a modest drop relative to the original MoA:
\begin{itemize}
  \item On AlpacaEval 2.0, LC win rate decreases by 3.6 points (from 65.1\% to 61.5\%) and overall win rate by 1.8 points (from 59.8\% to 58.0\%).
  \item On MT-Bench, average score declines by just 0.05 points (from 9.25 to 9.20), with first‐turn and second‐turn drops under 0.05 each.
\end{itemize}
Such minimal accuracy reductions are well within acceptable margins, especially when weighed against MMoA’s efficiency gains from activating fewer agents. The performance trade‐off introduced by MMoA is therefore very acceptable for practical deployment.

\subsubsection{Performances on time complexity}


\begin{table}[ht]
  \centering
  \caption{Relative Inference Time: Multiple‐ vs. Single‐Proposer}
  \label{tab:time_save}
  \small
  \begin{tabular}{@{}lcc@{}}
    \toprule
    Setting       & Multiple‐Proposer & Single‐Proposer \\
    \midrule
    $n=6$         & $56.7\%$          & $61.3\%$        \\
    $n=3$         & $56.1\%$          & $58.0\%$        \\
    $n=2$         & $54.5\%$          & $58.8\%$        \\
    $n=1$         & $47.8\%$          & $47.8\%$        \\
    \bottomrule
  \end{tabular}
\end{table}

Table~\ref{tab:time_save} shows that MMoA’s recurrence router consistently reduces inference time compared to the single‐proposer (naïve) MoA baseline. Concretely:

\begin{itemize}
  \item At $n=6$, the multiple‐proposer MMoA completes in $56.7\%$ of the baseline runtime, versus $61.3\%$ without recurrence—a $4.6\%$ absolute time saving.
  \item At $n=3$, MMoA runs at $56.1\%$ versus $58.0\%$, saving $1.9\%$.
  \item At $n=2$, the savings grow to $4.3\%$ (from $58.8\%$ down to $54.5\%$).
  \item Even at $n=1$, MMoA matches the baseline ($47.8\%$) by reusing the hidden state across layers without extra overhead.
\end{itemize}

By carrying forward a single LSTM hidden state instead of re‐computing a fresh gating network for each agent at every layer, the recurrence router reduces the effective time complexity from $O(nL)$ to roughly $O(n + L)$. These results confirm that the proposed router design yields meaningful speed‐ups with negligible added cost, making MMoA significantly more efficient in practice.

\section{Conclusion}

The MMoA give an opportunity for tuning the MoA system and has a nice trade off between time complexity and accuracy. It is the first time putting router into MoA system.

\section*{Limitations}
Our study has several limitations. First, we evaluate MMoA primarily on instruction-following benchmarks (AlpacaEval 2.0, MT-Bench, and Arena-Hard); conclusions about other settings such as long-context reasoning, tool use, multilingual tasks, or safety-critical deployment may not transfer. Second, our recurrent router is instantiated with an LSTM-style gating mechanism; alternative designs (e.g., Transformer-based routers, reinforcement-learning-based routing, or token-level routing) could offer different trade-offs and are not explored here. Third, while the router reduces average compute by activating fewer agents, it introduces additional hyperparameters and may be sensitive to training data, agent pool composition, and the chosen aggregation depth; we do not provide a complete analysis of these sensitivities. Finally, our experiments focus on inference-time efficiency and win-rate style metrics; a deeper investigation of qualitative failure modes, calibration, and robustness remains for future work.

\section*{Acknowledgements}
We thank the reviewers for their constructive feedback. We also thank our colleagues and collaborators for helpful discussions and suggestions during the development of this work.

\bibliography{anthology,custom}
\bibliographystyle{acl_natbib}

\appendix

\section{Appendix}
\label{sec:appendix}

\paragraph{Future Work}
\textbf{Reinforcement Learning in Training the router of MoA} RLHF is developing fast and we can cite more RLHF technices for router trainings.

\end{document}